\documentclass[runningheads]{llncs}
\usepackage[utf8]{inputenc}
\usepackage{amsmath}
\usepackage{amssymb}	
\usepackage{amsfonts}
\usepackage{dsfont}
\usepackage{hyperref}
\usepackage{graphicx}
\usepackage[dvipsnames]{xcolor}
\usepackage[linesnumbered,lined,boxed,commentsnumbered]{algorithm2e}
\usepackage[normalem]{ulem}

\usepackage{my_macros}

\begin{document}
\title{Metamorphic image registration using a semi-Lagrangian scheme}
%
%
\author{Anton François\inst{1,2}
\and
Pietro Gori\inst{2}
\and
Joan Glaunès\inst{1}
} 
\authorrunning{A. François et al.}
%
\institute{MAP5, Université de Paris, France\footnote{\url{https://math-info.u-paris.fr/en/map5-laboratory/}}
\email{\{anton.francois,alexis.glaunes\}@parisdescartes.fr}\\\and
LTCI, Télécom Paris, IPParis, France
\email{pietro.gori@telecom-paris.fr}\\}
\maketitle              
\begin{abstract}
In this paper, we propose an implementation of both \emph{Large Deformation Diffeomorphic Metric Mapping} (LDDMM) and Metamorphosis image registration using a semi-Lagrangian scheme for geodesic shooting. We propose to solve both problems as an inexact matching providing a single and unifying cost function.  We demonstrate that for image registration the use of a semi-Lagrangian scheme is more stable than a standard Eulerian scheme. Our GPU implementation is based on \texttt{PyTorch}, which greatly simplifies and accelerates the computations thanks to its powerful automatic differentiation engine. It will be freely available at \url{https://github.com/antonfrancois/Demeter_metamorphosis}.

\end{abstract}

\section{Introduction}

    
Diffeomophic image matching is a key component in computational anatomy for assessing morphological changes in a variety of cases. Since it does not modify the spatial organization of the image (i.e. no tearing, shearing, holes), it produces anatomically plausible transformations. Possible applications are: alignment of multi-modal images, longitudinal image registration (images of the same subject at different time points) or alignment of images of the same modality across subjects, for statistical analysis such as atlas construction. 

Extensive work has been conducted to efficiently compute diffeomorphic transformations. 
One strategy is to use flows of time-dependent vector fields, as in LDDMM \cite{Beg2005,Vialard2011,Ashburner2011}. This allows the definition of a right-invariant  metric on the group of diffeomorphisms as well as a Riemannian metric on the space of topologically consistent images,
which thus becomes a shape space, 
providing useful notions of geodesics, shortest paths and distances between images \cite{Dupuis1998,Younes2010}. A shortest path represents the registration between two images.
Due to the high computational cost of LDDMM, some authors proposed to use flows of stationary vector fields, instead than time-varying ones, with the Lie algebra vector field exponential \cite{arsigny_2006,Ashburner2007}. 

Diffeomorphic maps are by definition one-to-one, which means that they are suited for matching only images characterized by the same topology. However, many clinical or morphometric studies often include an alignment step between a healthy template (or atlas) and images with lesions, alterations or pathologies, like white matter multiple sclerosis or brain tumors. Three main strategies have been proposed in the literature. Cost function masking is used in order not to take into account the lesion or tumor (by masking it) during registration \cite{Ripolls2012}. This method is quite simple and easy to implement but it does not give good results when working with big lesions or tumors. The other two strategies consist in modifying either the healthy template or the pathological image in order to make them look like a pathological or healthy image respectively.
For instance, in GLISTR \cite{Gooya2012}, authors first make growing a tumor into an healthy image and then they register it to an image with tumor. This strategy is quite slow and computationally heavy. On the other hand, in \cite{sdika_nonrigid_2009}, authors try to fill the lesions using inpainting in order to make the image looks like an healthy image. This strategy seems to work well only with small lesions. Similarly, in \cite{liu_low-rank_2015}, authors proposed to estimate the healthy version of an image as its low-rank component, which seems to work correctly only when the sample size is quite large.  
Here, we propose to use a natural extension of LDDMM: the Metamorphosis framework as introduced in \cite{Trouv2005,Holm2009,Younes2010}. It was designed to jointly estimate a diffeomorphic deformation and a variation in image appearance, modeling, for instance, the apparition of a tumor. Intensity variations are also deformed along the images during the process. Metamorphosis is related to morphing in computer graphics and, like LDDMM, generates a distance between objects. However, in Metamorphosis the integration follows two curves, one for the deformation and one for the intensity changes (see definition in \cite{Holm2009,Younes2010,Gris_2020}). By comparing the works of Beg and Vialard \cite{Beg2005,Vialard2011} for LDDMM and Metamorphosis, one can notice that they are closely related as they end up using a very similar set of variational equations for their geodesics. 

We decided to use geodesic shooting \cite{Miller2006} as it is the only method that can theoretically ensure to get optimal paths (geodesics), even if not performing optimization until convergence.
Once defined the Euler-Lagrange equations associated to the functional of the registration, one can integrate them. The estimated optimal paths, which are geodesics, are usually computed using the shooting algorithm and are completely encoded by the initial conditions of the system. The minimizing initial conditions, subject to the geodesic equations, are usually estimated using a gradient descent scheme (which needs the computation of the adjoint equations).

In this paper, we make the following contributions. To the best of out knowledge, we provide the first implementation of both LDDMM and Metamorphosis joined in one optimisation problem.
We also propose a full semi-Lagrangian scheme for the geodesic shooting equations and
we give access to our easy to use GPU implementation fully developed with \texttt{PyTorch}.

\section{Methods}

\subsubsection{LDDMM and Metamorphoses geodesics.}

Let $\Omega \subset \R^d$ be a fixed bounded domain, where $d=\{2,3\}$. We define a gray-scale image $I \in \manim$ (or a gray-scale volume) as a square-integrable function defined on $\Omega$ (i.e.: $\manim \doteq L^2(\Omega,\R)$). 
Let $V$ be a fixed Hilbert space of $l$-times continuously differentiable vector fields supported on $\Omega$ 
(i.e.: $V \subset  \mathcal C ^l_0(\Omega,\R^d)$). We consider that the time varying vector fields $v_t$ are elements of $L^2(\left[0,1 \right];V),t\in[0,1]$, \cite{Beg2005,Younes2010,Vialard2011}. A pure deformation flow can be deduced by solving the ODE $\dot \Phi_t = v_t \cdot \Phi_t :=v_t \circ \Phi_t$ , $v_t \in V, \forall t \in [0,1]$.
Hence, the flow at a given time $t$ is written as $\Phi_t = \int_0^t v_s\circ \Phi_s ds$ with $\Phi_0=\mathrm{Id}$. As shown in  \cite{Dupuis1998}, the elements of $V$ need to be sufficiently smooth to produce a flow of diffeomorphisms. The traditional optimisation problem for LDDMM is an inexact matching. Let $\si, \ti \in \manim$ be a source and a target image, it aims at minimizing a cost composed of a data term
(e.g. the $L_2$-norm, known as the sum of images squared difference (SSD) ),
and a regularisation term on $v$, usually defined as the total kinetic energy  $\int_0^1 \|v_t\|_V^2 \ dt $. The goal is thus to find the "simplest" deformation to correctly match $\si$ to $\ti$.

Metamorphoses join additive intensity changes with the deformations.
The goal of Metamorphosis is to register an image $\si$ to $\ti$ using variational methods with an intensity additive term $z_t \in L^1(\left[0,1 \right];V)$.
The image evolution can be defined as :

\begin{equation}
    \partial_t\im{t} = v_t\cdot \im{t}+ \mu z_t = - \left< \nabla I_t,v_t\right> + \mu z_t, \quad \text{ s.t. } I_0=I \quad \mu\in \R^+.
    \label{eq_imevol}
\end{equation}

One can control the amount of deformation vs photometric changes by varying the hyperparameter $\mu \in \R^+$. The dot notation is used for infinitesimal step composition, here writing $v \cdot \im{t}$ implies that $\im{t}$ is deformed by an infinitesimal vector field $v$. 
As described by Trouvé \& Younes in \cite{Trouv2005,Younes2010}, the $\{ z_t \}$ have to be the 'leftovers' of the transport of $I_t$ by $v_t$ toward the exact registration, as it can be seen by rewriting Eq.\ref{eq_imevol} as $z_t = \frac{1}{\mu^2}(\partial_t I_t - v_t \cdot I_t)$. This is usually called the Metamorphic residual image or the momentum. 

In order to find the optimal $(v_t)_{t \in [0,1]}$ and $(z_t)_{t \in [0,1]}$, one can minimize the exact matching functional \cite{Younes2010,Holm2009,richardson_metamorphosis_2016} using Eq.\ref{eq_imevol}:
\begin{equation}
    E_M(v,z) = \int_0^1\|v_t\|^2_V + \rho \|z_t\|^2_{L_2} dt, \quad \text{s.t. } I_1=J ,  I_0=I;\quad \rho \in \R^+,
    \label{eq_costM}
\end{equation}

As shown in \cite{Younes2010,Holm2009}, the geodesic equations for Metamorphosis are:
\begin{equation}
		\left\{
		\begin{array}{rl}
			v_t &= - \frac\rho\mu K \star  (z_t \nabla I_t)\\
			\partial_t z_t &= -\quad \nabla \cdot (z_t v_t)  \\
			\partial_t I_t &= - \left< \nabla I_t,v_t\right> + \mu z_t
		\end{array}
		\right.
		\label{eq:geodesic}
\end{equation}

By setting $\rho=\mu$ and letting $\mu\rightarrow 0$, one recovers the geodesic equations for LDDMM as pointed out in \cite{Younes2010,Vialard2011}. 
In Eq. \ref{eq:geodesic},   $\nabla \cdot (zv) = \textrm{div}(zv)$ is the divergence of the field $v$ times $z$ at each pixel, $K $ is the chosen translation invariant reproducing kernel (of the RKHS) and $\star$ is the convolution. In practice $K$ is often a Gaussian blurring kernel \cite{Miller2006,Vialard2011}. 
The last line of Eq. \ref{eq:geodesic} is the advection term, simulating the movement of non diffusive material. 
The second (continuity) equation is a conservative form which ensures that the amount of deformation is preserved on the whole domain over time. Thus, given the initial conditions of the system, $I=I_0$ and $z_0$, one can integrate in time the system of Eqs. \ref{eq:geodesic} to obtain $I_1$. Note that $v_0$ can be computed from $z_0$, making $z$ the only unknown.  
Furthermore, one can notice that the energy in Eq.\ref{eq_costM} is conserved (i.e.: constant along the geodesic paths) and therefore the time integrals may be replaced by the norms at time 0.

Here, we propose to solve Metamorphosis as an inexact matching problem. This allows us to have a unifying cost function (i.e.: Hamiltonian) for both LDDMM and Metamorphosis:
\begin{equation}
    H(z_0) =   \frac 12\| \im{1} - \ti \|_{L_2}^2 + \lambda \Big[ \|v_0\|^2_V + \rho  \|z_0\|^2_{L_2} \Big]
\label{eq_cost}
\end{equation}
with $\|v_0\|_V^2 = \left< z_0 \nabla \si, K \star (z_0 \nabla \si)\right>$ . 
The hyperparameters $\lambda$ and $\rho$ define the amount of total regularization and intensity changes respectively. 

\subsubsection{Geodesic shooting integration}
Integration of the geodesics is a crucial computational point for both LDDMM and Metamorphosis. In the case of image registration using LDDMM, Beg et al. \cite{Beg2005} initially described a method based on gradient descent which could not retrieve exact geodesics, as shown in \cite{Vialard2011}.  An actual shooting method was then proposed in \cite{Vialard2011} for LDDMM based registration of images. To the best of our knowledge, the only shooting method proposed in the literature for image Metamorphosis is the one proposed in \cite{richardson_metamorphosis_2016}. It is based on a Lagrangian frame of reference and therefore it is not well suited for large images showing complicated deformations, as it could be the case when registering healthy templates to patients with large tumors. Here, we propose to use a semi-Lagrangian scheme.


    
\subsubsection{From Eulerian to semi-Lagrangian formulation \label{seq:eul2Lag}}

When analysing flows from ODE and PDE, two concurrent points of view are often discussed. Lagrangian schemes where one follows the stream of a set of points from initialisation, and Eulerian schemes where one consider some fixed point in space (often a grid) and evaluate the changes occurred. Eulerian schemes seem to be the most natural candidate for flow integration over an image. In fact, in the Lagrangian schemes the streams of pixels we follow can go far apart during integration, making impossible the image re-interpolation. However, the Lagrangian scheme is more numerically stable than the Eulerian one thus allowing a better convergence \cite{avants_lagrangian_2006}. 
Indeed, a necessary condition for proper convergence for Eulerian schemes is the Courant–Friedrichs–Lewy(CFL) condition, which defines a maximum value for the time step. If the time step is badly chosen, ripples may appear or, worse, the integration may fail (see Fig.\ref{fig_eulvsl}). The minimal number of time steps required increases with the size of the image, thus increasing the number of iterations and making it very slow to use on real imaging data. 
For these reasons, the so-called semi-Lagrangian scheme seems to be a good compromise.
The idea is to compute the deformation of a grid corresponding to a small displacement $\mathrm{Id}-\delta t\, v_t$, and then interpolate the values of the image $I_t$ on the grid. This can be summarized by $I_{t+\delta t}\approx I_t \circ (\mathrm{Id}-\delta t\, v_t)$. Semi-Lagrangian schemes are stable and don't need many iterations. Too many iterations would blur the images due to the successive  bilinear or trilinear (in 3D) interpolations. 

Let's reformulate Eq.\ref{eq:geodesic} in a semi-Lagrangian formulation, starting by the advection part \cite{Efremov2014}. From the Eulerian formulation for the closed domain $[0,1] \times \Omega$, we can write:
\begin{equation}
    \partial_t \im{t} + \left< \nabla \im{t},v_t\right> - \mu z_t = \partial_t \im{t} + \sum_{i=1}^d \partial_{x_i} I(t,x) v_{x_i}(t,x) - \mu z(t,x) = 0.
    \label{eq_semi_advec}
\end{equation}
where we use for convenience the notations $v_t(x) = v(t,x)= (v_{x_1}(t,x),\cdots,v_{x_d}(t,x)),t\in[0,1],x \in \Omega$. We deduce the characteristics defined by the system of differential equations :
\begin{equation}
x_i'(t) = v_{x_i}(t,x) + \mu z_t, \qquad i \leq d \in \N^*, t \in [0,1].
\end{equation}
Then, we can also rewrite the continuity equation as:
\begin{align}
    \partial_t z_t + \nabla \cdot (z_t v_t) &= \partial_t z_t + \sum_{i=1}^d \partial_{x_i}(z(t,x) \times v_{x_i}(t,x)) =0\\
     &= \partial_t z_t + < \nabla z_t, v_t > + (\nabla \cdot v_t)z_t =0
\end{align}
In the same way, we also extract the characteristics defined by the system of differential equations :
\begin{equation}
x_i'(t) = v_{x_i}(t,x) + (\nabla \cdot v(t,x)z(t,x), \qquad i \leq d \in \N^*, t \in [0,1].
\label{eq_semi_ad_z}
\end{equation}
Note that by using a semi-Lagrangian scheme for this part we avoid to compute a discrete approximation of $\nabla \cdot (z_t v_t)$, but still need to compute an approximation of $\nabla \cdot v_t$. However, the momentum $z_t$, similarly to the image $I_t$, is potentially non smooth, while $v_t$ is smooth due to its expression through the convolution operator $K$.

In the following Section, we will compare three computational options to integrate over the geodesics: 1- the Eulerian scheme, 2- the semi-Lagrangian approach and 3- a combination of the two, where we use the semi-Lagragian scheme for the advection (Eq.\ref{eq_semi_advec}) and the Eulerian scheme for the residuals (Eq. \ref{eq_semi_ad_z}).
    

\section{Results and conclusions}

\begin{figure}
\centering
    \includegraphics[width=\textwidth]{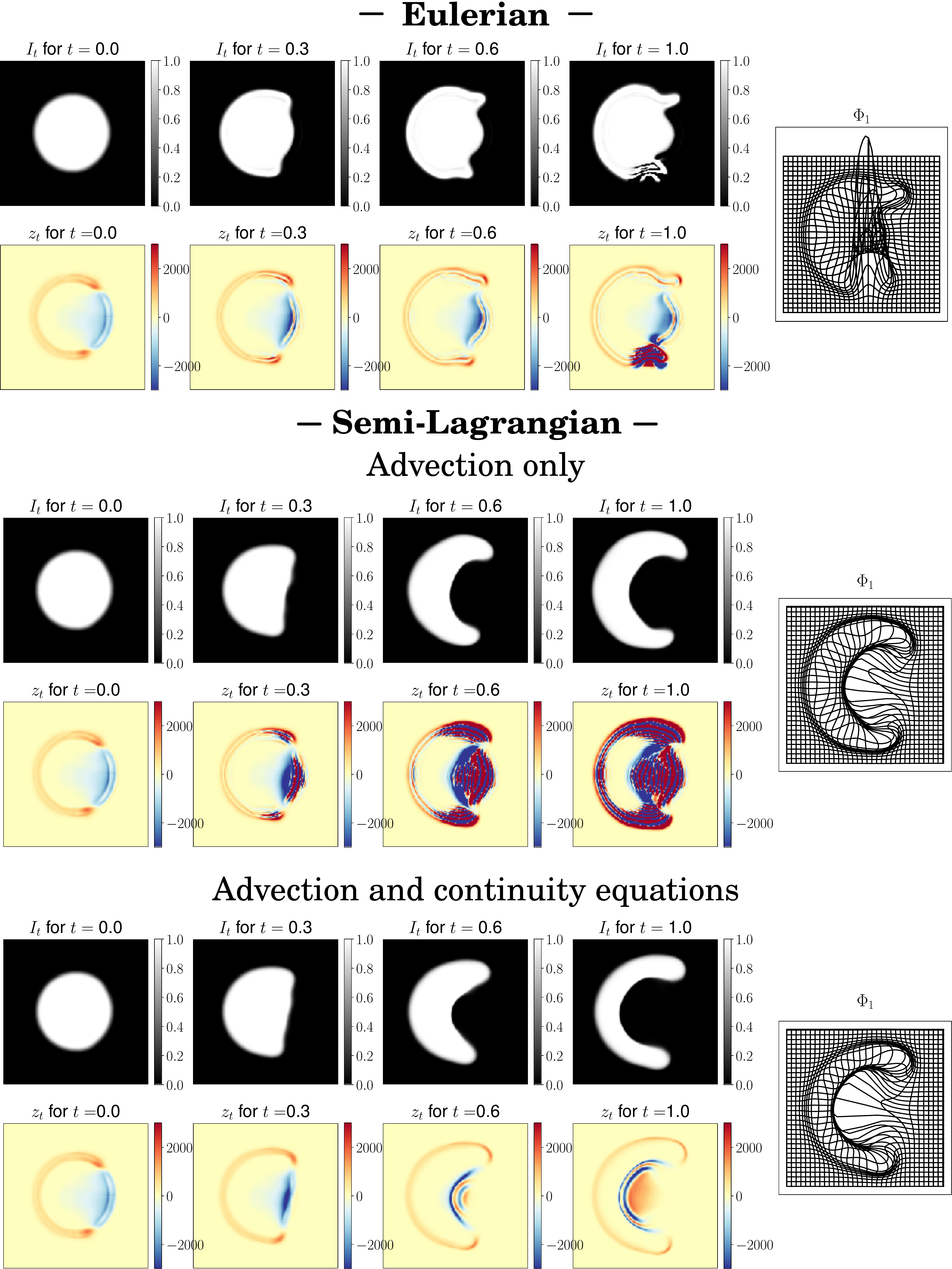}
    \caption{\textbf{Comparison between the stability of the 3 geodesic shootings schemes proposed for LDDMM}
    In each row, we show four intermediary shooting steps with the same initial $z_0$ and RKHS for $v$.
    The black and white pictures are the images, with below the corresponding $z$. The deformations grids on the right are obtained by integrating over all $v_t, \forall t\in [0,1] $. Shooting was performed using a $z_0$ obtained from LDDMM optimisation towards a 'C' picture ($\mu =0$). The Eulerian and semi-Lagrangian schemes have a time step of 1/38 and 1/20 respectively.
    \label{fig_eulvsl}}
\end{figure}

\begin{figure}[t]
\centering
    \includegraphics[width=\textwidth]{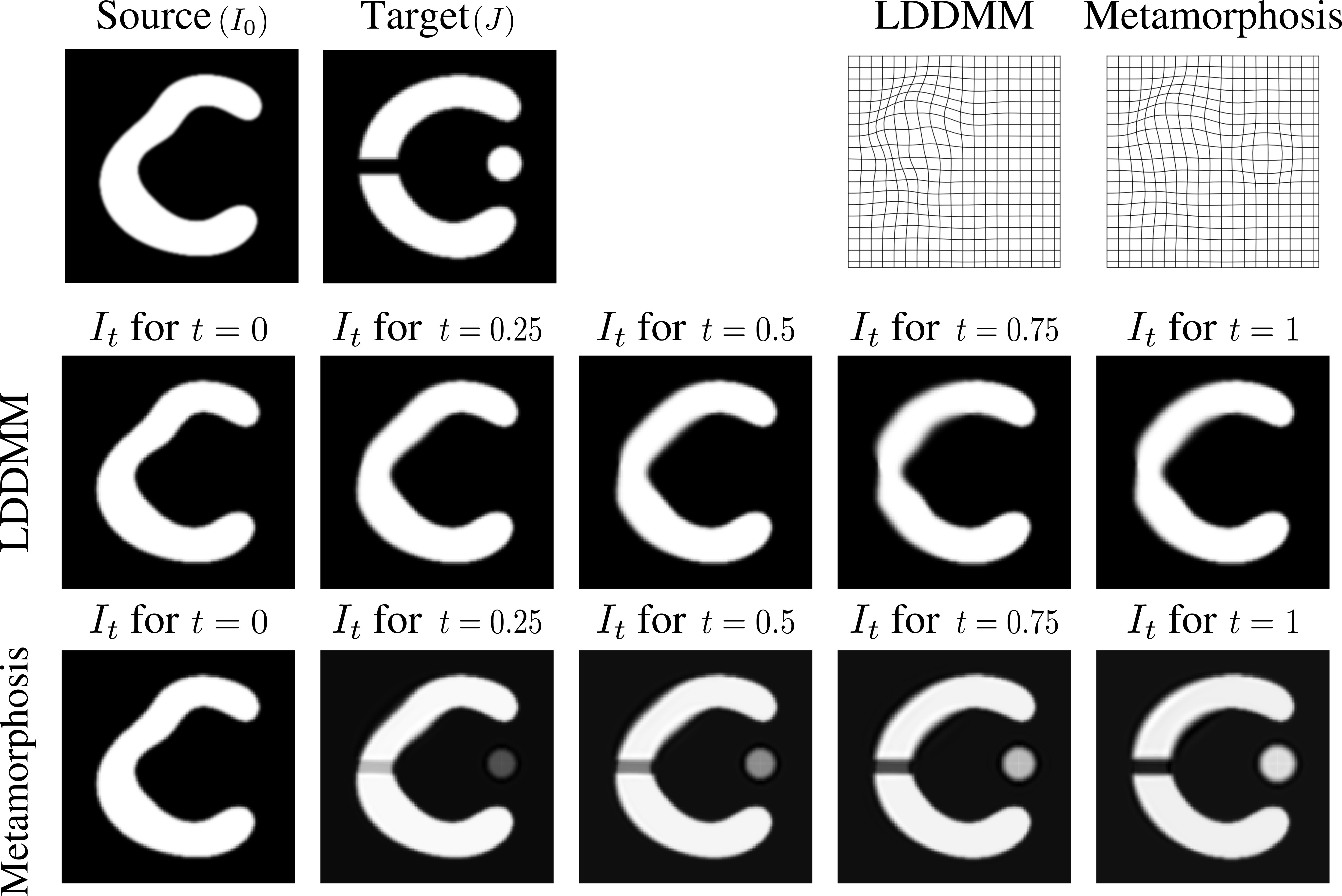}

   \caption{\textbf{Comparison of LDDMM vs Metamorphoses registration} \emph{Top-right}  final deformation grids 
obtained by integrating over all vector fields $(v_t)_{t \in [0,1]}$.
     \emph{Bottom Rows} Image evolution during the geodesic shootings of the respective method after optimisation with Eq. \ref{eq_cost}. \label{fig_metas}}
\end{figure}

In Fig. \ref{fig_eulvsl}, we can observe the lack of stability of Eulerian methods compared to the semi-Lagrangian ones. Even if the chosen time step is rather small, the Eulerian scheme produces ripples (in purple in the residuals) and the integration fails (see the estimated deformation). On the contrary, semi-Lagrangian schemes converge to a better deformation with an higher time step. It should also be noted that the full semi-Lagrangian scheme (advection and continuity equations) is perfectly stable without showing ripples, as it is instead the case for the advection-only semi-Lagrangian scheme.

In Fig. \ref{fig_metas} we can see that Metamorphosis and LDDMM describe deformations in a similar way. As we use the SSD (i.e. $L^2$ norm) as data term, an object in the source image is matched to another object in the target image only if they have some pixels in common. For this reason, the C form is not pushed to match the small disk on the right of the example. In the Metamorphosis deformation grid we can see that the small disk is growing, as it is less costly to create a small disk and make it grow.
However, the Metamorphic registration, thanks to the intensities changes modeled by $z$, manages to correctly take into account the topological differences between the source and target images.

With the use of automatic differentiation, we bypass the extensive and delicate work of deriving the backward adjoint equations and  implementing a discrete scheme to solve them.
This allowed us to merge LDDMM and Metamorphosis into a single framework and to easily test different configurations of the problem.
For this study, we optimized all costs using gradient descent. We also provide alternative optimization methods, such as L-BFGS, in our library Demeter, which will be regularly updated.

\bibliographystyle{splncs04}
\bibliography{carnetDeRoute}


%
%
%
%

\end{document}